\documentclass{article}

\usepackage[preprint]{corl_2026} 

\usepackage[utf8]{inputenc}
\usepackage[T1]{fontenc}
\usepackage{url}
\usepackage{booktabs}
\usepackage{amsfonts}
\usepackage{amsmath}
\usepackage{amssymb}
\usepackage{mathrsfs}
\usepackage{bm}
\usepackage{nicefrac}
\usepackage{microtype}
\usepackage[table]{xcolor} 
\usepackage{graphicx}
\usepackage{bbding}
\usepackage{multirow}
\usepackage{multicol}
\usepackage{makecell}
\usepackage{wrapfig}
\usepackage{float}
\usepackage{subcaption}
\usepackage{array}
\usepackage{caption}
\usepackage{enumitem}
\usepackage[ruled,vlined]{algorithm2e}

\title{SkillMemo: Expert-guided Skill Memory Framework for Compositional Embodied Manipulation}

\author{%
  {\normalfont Changyuan Wang\textsuperscript{1}\hspace{0.15cm}
  Chubin Zhang\textsuperscript{1}\hspace{0.15cm}
  Zhenyu Wu\textsuperscript{2}\hspace{0.15cm}
  Runhao Li\textsuperscript{3}\hspace{0.15cm}
  Angyuan Ma\textsuperscript{2}\hspace{0.15cm}
  Ke Chao\textsuperscript{4}} \\[1.5mm]
  {\normalfont Yinan Liang\textsuperscript{2}\hspace{0.15cm}
  Xiuwei Xu\textsuperscript{2}\hspace{0.15cm}
  Ziwei Wang\textsuperscript{3}\hspace{0.15cm}
  Yansong Tang\textsuperscript{1}\thanks{Corresponding author.}\hspace{0.15cm}
  Jiwen Lu\textsuperscript{2}} \\
  \\
  \textsuperscript{1}Shenzhen International Graduate School, Tsinghua University \\
  \textsuperscript{2}Department of Automation, Tsinghua University \\
  \textsuperscript{3}Nanyang Technological University \hspace{0.5cm}
  \textsuperscript{4}Beijing Normal University \\[1.2mm]
  \textbf{Project Page:} \url{https://changyuanwang17.github.io/SkillMemo/}
}

\begin{document}
\maketitle
\vspace{-1.2mm}

\begin{abstract}
Embodied visuomotor models, including Diffusion Policy (DP) and Vision-Language-Action (VLA) models, have demonstrated promising performance on robotic manipulation benchmarks. However, their potential remains fundamentally constrained by the scarcity of large-scale embodied trajectory datasets, leading to insufficient compositional generalization in out-of-distribution (OOD) scenarios with limited capability to capture reusable skill structures. To address this limitation, we propose Skill-Based Memory (SkillMemo) framework that implicitly decomposes long-horizon demonstrations into latent atomic skills and integrates skill-level features into a dynamic episodic memory bank for solving compositional tasks. Specifically, we first introduce an expert-guided trajectory segmentation module built upon a Mixture-of-Experts (MoE) architecture, which implicitly partitions trajectories into distinct skill primitives represented by learned gating coefficients. We further design a skill-level episodic memory architecture that stores compact skill representations as retrievable key-value pairs. During inference, the memory bank retrieves the most relevant skill primitives which are subsequently fused with the model's current gating distribution, providing a robust contextual prior to refine action predictions. Extensive experiments on the simulation benchmark and real-world manipulation tasks demonstrate that SkillMemo consistently enhances both DP and VLA backbones, achieving state-of-the-art performance and outperforming $\pi_{0.5}$, while exhibiting strong compositional generalization to unseen task configurations.
\end{abstract}



\section{Introduction}

Embodied visuomotor models, including Diffusion Policy (DP)~\citep{chi2023diffusion, wang2024sparse, rana2025imle} and Vision-Language-Action (VLA)~\citep{qu2025spatialvla, deng2025graspvla, bu2025univla, intelligence2025pi_, kim2024openvla} models, have achieved remarkable success in robotic manipulation, primarily driven by large-scale vision-language pretraining followed by embodied trajectory finetuning~\citep{o2024open, khazatsky2024droid, bu2025agibot}. Nonetheless, their real-world deployment remains fundamentally limited by the high cost of acquiring large-scale robot datasets. Training on insufficient demonstrations often results in monolithic policies that fail to capture reusable behavioral structures, leading to degraded performance when facing out-of-distribution (OOD) scenarios wherein novel compositions of known objects, receptacles, and instructions.

To address these challenges, prior efforts have investigated robotic reinforcement learning~\citep{luo2025precise, wagenmaker2025steering, ankile2025residual}, motion primitive libraries~\citep{zheng2025universal, yao2025think, mao2024robomatrix}, and memory-augmented policies~\citep{shi2025memoryvla, koo2025hamlet}. Among these, memory-augmented methods have emerged as a promising paradigm for long-horizon generalization. MemoryVLA~\citep{shi2025memoryvla} introduces a dual-stream Perceptual-Cognitive Memory Bank for context-aware decision making. HAMLET~\citep{koo2025hamlet} addresses non-Markovian dependencies by encoding interaction history into temporal moment tokens.
However, existing methods lack an explicit mechanism for memorizing behaviorally salient skill-level features. Consequently, they tend to exhibit inefficient reuse of prior experience and diminished capability for compositional generalization in unseen tasks.

In this paper, we propose a Skill-Based Memory (SkillMemo) framework to enhance robust compositional generalization for embodied manipulation. Unlike conventional visuomotor models that encode entire behavioral trajectories with a single monolithic network, SkillMemo implicitly decomposes long-horizon demonstrations into latent atomic skills and integrates skill-level features into a dynamic episodic memory bank for solving compositional tasks. Specifically, we first introduce an expert-guided trajectory segmentation module built on a Mixture-of-Experts (MoE) architecture, which implicitly partitions trajectories into distinct skill primitives represented by learned gating coefficients. Building upon this segmentation, we design a skill-level episodic memory architecture that stores compact skill representations as retrievable key-value pairs. During inference, the memory bank retrieves the most relevant skill primitives which are subsequently fused with the model's current gating distribution, providing a robust contextual prior to refine action predictions. Extensive experiments on the simulation benchmark and real-world UR5e manipulation tasks demonstrate that SkillMemo consistently improves both DP and VLA backbones, achieving state-of-the-art performance and outperforming $\pi_{0.5}$ by 1.2\% (98.0\% vs.\ 96.8\%), while exhibiting strong compositional generalization to unseen task configurations.


\begin{figure*}[t]
	\centering
        \includegraphics[width=0.98\linewidth]{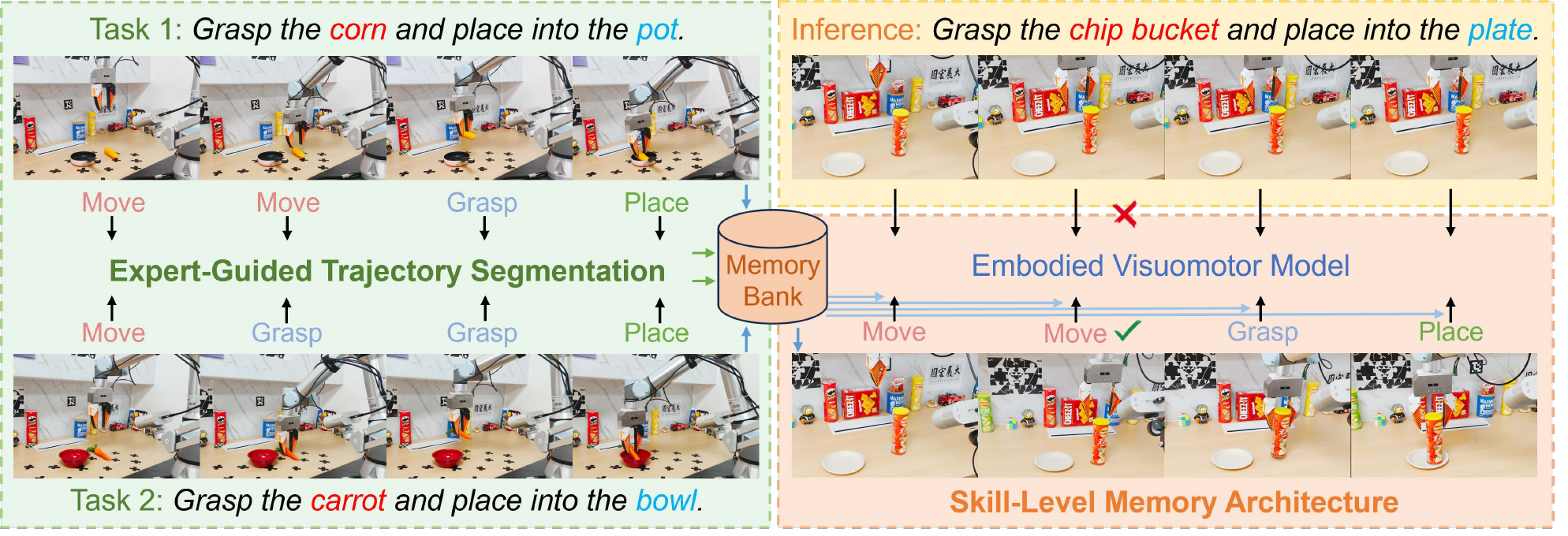}
        \vspace{-0.2cm} 
	\caption{Unlike conventional models that struggle to capture reusable skill structures, our method implicitly decomposes trajectories into latent atomic primitives for memory storage, which are dynamically retrieved and fused to facilitate compositional generalization to unseen task configurations.}
	\vspace{-0.32cm}     
	\label{comparison}
\end{figure*}




\section{Related Works}

\subsection{Embodied Visuomotor Model}
Recent advances in embodied AI have been largely driven by Diffusion Policy (DP)~\citep{chi2023diffusion} and Vision-Language-Action (VLA) models~\citep{zitkovich2023rt, kim2024openvla, black2024pi_0}. While DP formulates control as iterative denoising and has been extended via 3D representations~\citep{ze20243d, ze2024generalizable} and hierarchical structures~\citep{lu2025h, wang2024sparse}, VLA models couple large-scale vision-language pretraining with action generation using extensive robot datasets~\citep{walke2023bridgedata, o2024open, bu2025agibot, intelligence2025pi_, li2025controlvla}. Despite their impressive zero-shot capabilities, both paradigms struggle with compositional generalization when limited demonstrations are available, motivating our work on skill-level memory augmentation.

\subsection{Skill Learning}
To address the temporal complexity of manipulation, recent approaches explore skill decomposition. Explicit frameworks~\citep{mao2024robomatrix, li2025atomic, yao2025think} decouple high-level planning from execution by constructing dynamic libraries of motor primitives or prompts for task composition. Alternatively, implicit methods discover skills within the policy architecture using phase-aware masking~\citep{fan2025long} or Mixture-of-Experts (MoE)~\citep{wang2024sparse, huang2024mentor}. Distinct from methods relying on static skill libraries or purely implicit partitioning, our approach uniquely couples MoE-based skill discovery with a dynamic episodic memory bank, enabling structured storage and compositional retrieval of learned skills.

\subsection{Memory-based Framework}
To mitigate the limitations of purely reactive policies, recent research integrates explicit memory mechanisms into visuomotor architectures. A prominent example is MemoryVLA~\citep{shi2025memoryvla}, which utilizes a dual-stream Perceptual-Cognitive Memory Bank to store visual details and semantic summaries for context-aware retrieval. While other works explore temporal tokens or learnable prompts~\citep{lei2025robomemory, li2025map, koo2025hamlet, lan2025experience, lei2025dynamic}, these methods typically store unstructured or holistic visual-semantic representations. They lack the capacity to capture and reuse fine-grained, skill-level behavioral patterns. In contrast, SkillMemo structurally stores decomposed skill primitives and retrieves them based on behavioral relevance, providing robust contextual priors for compositional generalization.

\section{Methods}


\begin{figure*}[t]
  \centering
  \includegraphics[width=0.98\linewidth]{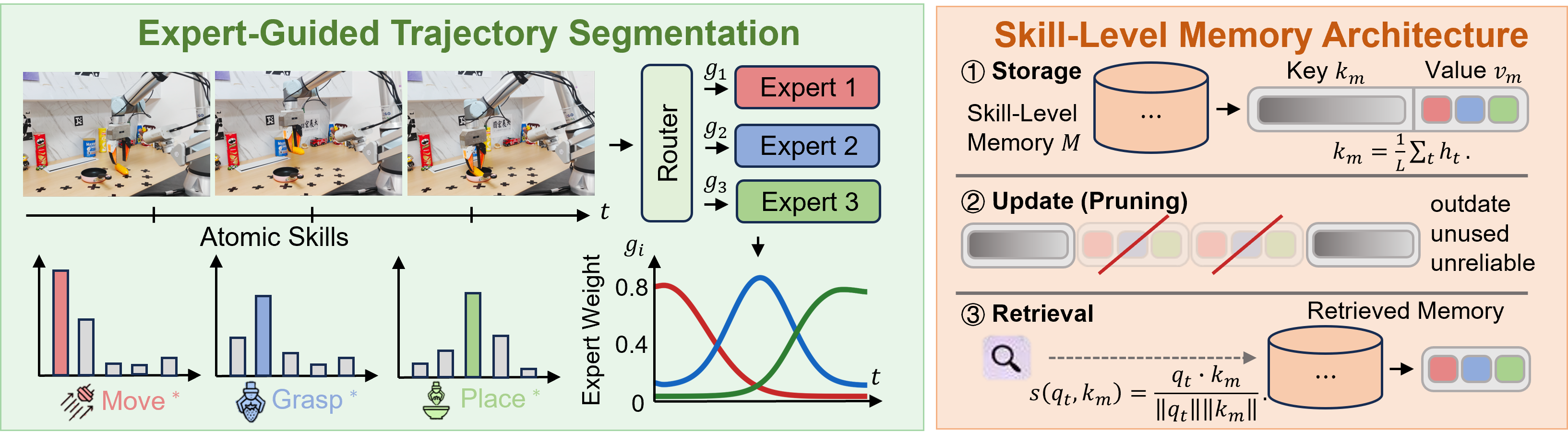}
  \vspace{-0.2cm}
  \caption{Overview of the SkillMemo framework. The Expert-Guided Trajectory Segmentation module implicitly$^*$ extracts latent atomic skills from demonstrations via MoE routing. The Skill-Level Memory Architecture structurally stores and dynamically prunes these learned gating profiles, enabling context-aware retrieval for robust compositional generalization on unseen task.}
  \vspace{-0.3cm}
  \label{fig:pipeline}
\end{figure*}

\subsection{Expert-Guided Trajectory Segmentation}
\label{sec:segmentation}

Conventional VLA models typically learn entire behavioral trajectories using a single monolithic network, which leads to limited compositionality and poor generalization over heterogeneous, high-dimensional action sequences. To address this limitation, we introduce an expert-guided trajectory segmentation module that structurally decomposes entire trajectories into a sequence of latent atomic skills. This decomposition enables fine-grained specialization of network capacity and facilitates structured skill reuse. Formally, given a full trajectory $\tau = \{ (x_t, a_t) \}_{t=1}^{T}$, where $x_t$ and $a_t$ denote the observation and action at time $t$, the explicit segmentation process partitions $\tau$ into a set of non-overlapping skill segments:
\begin{equation}
\tau \rightarrow \{ s_k \}_{k=1}^{K}, \quad s_k = \{ (x_t, a_t) \mid t \in [t^{\text{start}}, t^{\text{start+L}}] \},
\end{equation}
where each segment $s_k$ corresponds to a semantically coherent skill with the chunk size $L$. However, explicit segmentation schemes often fail to capture all potential skill boundaries which leads to inaccurate skill partitioning. On the contrary, we introduce an implicit skill partitioning scheme, where the decomposition is dynamically learned from the model's feature representations. Specifically, we implement this skill decomposition using a Mixture-of-Experts (MoE) architecture, which consists of $N$ expert networks $\{E_1, \ldots, E_N\}$ and a gating network $G$. Given an intermediate feature representation $h_t$ at time $t$, the MoE layer computes:
\begin{equation}
    y_t = \sum_{i=1}^N g_i(h_t) \cdot E_i(h_t),
\end{equation}
where $E_i(\cdot)$ denotes the $i$-th expert and $g_i(h_t)$ represents its corresponding gating weight. Intuitively, each expert captures a recurring local motion primitive, while the gating network $g$ implicitly defines the temporal segmentation boundaries by modulating expert activations over time. This implicit partitioning mechanism enables adaptive and data-driven skill discovery, ensuring that the identified skills align with emergent task-level structures.

To further encourage distinct skill specialization among experts and promote emergent skill composition, we regularize the model using a synergy-based information loss derived from the theory of \textit{Partial Information Decomposition (PID)}. For any pair of experts $(A,B)$, we define their synergistic information with respect to the task target ${G}$ as:
\begin{equation}
    \mathcal{L}_{\text{PID}} = -\big[I(\{A,B\};{G}) - I(A;{G}) - I(B;{G})\big],
\end{equation}
where $I(\cdot;\cdot)$ denotes mutual information. Minimizing $\mathcal{L}_{\text{PID}}$ encourages the joint representation of the experts to contain more task-relevant information than the sum of individual expert representations. This drives each expert to specialize in complementary skills while enabling higher-order coordination, thereby facilitating the learning of distinct, non-overlapping skill primitives. Complementarily, we employ expert load-balancing losses and top-k routing strategies to prevent expert collapse.

\subsection{Skill-Level Memory Architecture}
\label{sec:memory}

While the MoE-based trajectory segmentation in Section~\ref{sec:segmentation} discovers a set of reusable atomic skills, a critical challenge remains: how to effectively leverage these skills for solving compositional novel tasks where previously seen objects are recombined in ways not seen during training. Conventional VLA policies are purely reactive, relying only on the current observation and their static network weights, thereby discarding the rich historical context embedded in past successful executions. To endow our agent with a form of experiential reasoning, we introduce a Skill-Level Episodic Memory Architecture that stores, retrieves, and composes previously learned skill primitives during inference.

\noindent\textbf{\textit{Memory Storage.}} 
We first establish the foundational principle of the memory module: explicitly retaining previously acquired skills to facilitate task completion through adaptive skill composition during inference. To achieve this, we organize the memory bank $\mathcal{M}$ as a collection of Key-Value pairs, comprising two core components:

\noindent(1) \textit{Memory Key.} 
For each atomic skill segment $s_k = \{ (x_t, a_t) \}_{t=t^{\text{start}}}^{t^{\text{start+L}}}$ discovered by the segmentation module, we construct a compact memory entry. To enable efficient retrieval, we utilize the latent trajectory features $h_t$ to determine the similarity between the inference observation and the stored skills. However, directly storing frame-wise features for all timesteps results in an unscalable memory bank. To address this, we employ temporal aggregation to extract a compact centroid feature for each skill segment:
\begin{equation}
    \mathbf{k}_m = \frac{1}{L} \sum_{t \in \tau_m} h_t,
\end{equation}
where $\tau_m$ denotes the $m$-th skill segment and $h_t$ represents the latent feature at time $t$. By storing this compact centroid, we preserve the most representative characteristics of the skill trajectory, ensuring the efficiency of the retrieval process.

\noindent(2) \textit{Memory Value.} The value component stores the functional knowledge required to reproduce the skill. Since the gating weights $g(h_t)$ govern expert activation and implicitly define the category of the skill, we designate the gating distribution as the core knowledge content. To ensure the preservation of detailed temporal dynamics and information validity, we store the complete sequence of gating weights across all time steps within the segment. The final memory bank $\mathcal{M}$ is thus organized as:
\begin{equation}
    \mathcal{M} = \left\{ \left(\mathbf{k}_m, \mathbf{v}_m\right) \mid \mathbf{v}_m = \{ g(h_t) \}_{t \in \tau_m} \right\}_{m=1}^{M},
\end{equation}
where $M$ is the total number of stored skill primitives, and $\mathbf{v}_m$ represents the $m$-th stored gating weights. This memory structure enables the model to efficiently retrieve the most relevant skills and significantly enhances the inference process through expert guidance.

\noindent\textit{\textbf{Memory Retrieval.}} 
Following the construction of the memory bank during training, we leverage the stored procedural knowledge to enhance policy execution during inference. Standard VLA models often struggle with generalization when facing out-of-distribution (OOD) scenarios due to the scarcity of diverse training samples. Our skill-based memory addresses this by providing a structured bank of reusable behaviors. 

First, we identify the most relevant historical skills by comparing the current state representation with the stored memory keys. Let $q_t$ denote the latent feature of the current observation at inference time. We compute the cosine similarity between $q_t$ and every key $\mathbf{k}_m$ in the memory bank $\mathcal{M}$:
\begin{equation}
    s(q_t, \mathbf{k}_m) = \frac{q_t \cdot \mathbf{k}_m}{\|q_t\| \|\mathbf{k}_m\|},
\end{equation}
where $s(\cdot, \cdot)$ denotes the similarity score. Based on these scores, we retrieve the top-$N$ memory entries $\mathcal{M}_{\text{top}} = \{(\mathbf{k}_n, \mathbf{v}_n)\}_{n=1}^N$ that exhibit the highest affinity with the current state.

To ensure robustness, we introduce a memory reliability check before integrating the retrieved information. We calculate the average similarity score of the retrieved set and compare it against a predefined confidence threshold $\delta$. If the retrieval is deemed reliable, we proceed to augment the current policy by fusing the retrieved expert activation profiles with the model's current predicted gating distribution. Specifically, we refine the gating weights $g(h_t)$ by averaging them with the retrieved gating profiles:
\begin{equation}
    g'(q_t) = \lambda g(q_t) + (1 - \lambda) \frac{1}{N} \sum_{n=1}^N \mathbf{v}_n[t'],
\end{equation}
where $\lambda$ is a hyperparameter controlling the strength of the memory intervention, and $\mathbf{v}_n[t']$ corresponds to the aligned timestep within the retrieved skill sequence. By explicitly incorporating these retrieved priors, the model can synthesize novel behaviors through the composition of known primitives, significantly improving compositional generalization on unseen task configurations.

\noindent\textit{\textbf{Memory Update.}} 
To adapt to lifelong learning scenarios where the agent continuously encounters new tasks, we design a dynamic memory update mechanism for finite-capacity memory bank. Since the memory bank has a finite capacity $M_{\text{max}}$, it is crucial to prune obsolete information to accommodate new experiences. We implement a usage-tracking method that monitors the retrieval history and validation status of each memory entry.

When the memory bank reaches its capacity limit, we employ a prioritized pruning strategy to discard low-utility information. Specifically, we target memory entries for removal based on three criteria: (1) memories with the most outdated timestamps, mimicking the human cognitive process of forgetting distant, non-reinforced details; (2) memories that have remained unused, i.e., not successfully retrieved; (3) memories that consistently fail to meet the reliability threshold $\delta$ during the retrieval phase, indicating low-quality or irrelevant entries. This dynamic pruning strategy balances the trade-off between memory diversity and memory quality, ensuring that the memory bank remains compact, relevant, and conducive to effective skill composition.


\section{Experiments}


\subsection{Implementation Details}
We evaluate our proposed SkillMemo framework on multimodal embodied manipulation tasks by instantiating both diffusion policy (DP) and vision-language-action (VLA) frameworks as backbones. Specifically, we adopt Diffusion Policy~\citep{chi2023diffusion} for low-level visuomotor control and UniACT-0.5B~\citep{zheng2025universal}, UniVLA-7B~\citep{bu2025univla} as representative large-scale VLA models. Further experimental setups, including hyperparameter configurations and hardware specifics, are detailed in Appendix~\ref{app:implementation}.



\subsection{Ablation Studies}
\noindent\textbf{Component-wise Analysis.}  To isolate the contributions of each component, we conduct ablation studies under both the DP and VLA frameworks, evaluating task success rate on the LIBERO benchmark~\citep{liu2023libero}. Starting from the backbone model, we progressively integrate the proposed \textit{Expert-Guided Trajectory Segmentation} (EGTS) and \textit{Skill-Level Memory Architecture} (SLMA). Table~\ref{ablation_method} reports the success rates across different configurations. Compared to the vanilla backbone, adding EGTS alone yields consistent improvements across all task suites, confirming the effectiveness of implicit skill decomposition. Further incorporating SLMA provides additional gains, particularly on tasks requiring compositional reasoning.

\begin{table*}[t]
    \small
    \renewcommand\arraystretch{1.15}
    
    \begin{minipage}[t]{0.50\textwidth}
        \centering
        \caption{Effect of different methods we proposed, reporting accuracy of DP and UniACT model.}
        \label{ablation_method}
        \vspace{-0.05cm}
        \resizebox{\linewidth}{!}{%
        \begin{tabular}{c|cc|ccc}
            \hline
            Model & EGTS & SLMA & Goal & Spatial & Object\\
            \hline
            \multirow{3}{*}{UniAct} & & & 68.7 & 72.1 & 75.7 \\
             & \Checkmark & & 71.1 & 78.6 & 79.3 \\
             & \Checkmark & \Checkmark & \textbf{73.4} & \textbf{80.2} & \textbf{79.8} \\
            \hline
            Model & EGTS & SLMA & Push-T & BlockPush & Kitchen\\
            \hline
            \multirow{3}{*}{DP} & & & 52.9 & 73.6 & 57.1 \\
             & \Checkmark & & 53.6 & 76.5 & 59.1 \\
             & \Checkmark & \Checkmark & \textbf{55.2} & \textbf{78.1} & \textbf{60.6} \\
            \hline
        \end{tabular}}
    \end{minipage}%
    \hfill
    \begin{minipage}[t]{0.47\textwidth}
        \centering
        \caption{Effect of expert number on UniAct success rates and inference time.}
        \label{ablation_expert}
        \vspace{-0.05cm}
        \resizebox{\linewidth}{!}{%
        \begin{tabular}{c|ccc|c}
            \hline
            \makecell{Expert \\ Number ($N$)} & Goal & Spatial & Object & \makecell{Inference \\ Time (s)} \\
            \hline
            1  & 68.7 & 72.1 & 75.7 & 3.20 \\
            2  & 69.7 & 75.6 & 77.9 & 2.78 \\
            3  & 70.3 & 76.3 & 78.5 & 3.60 \\
            5  & \textbf{71.1} & 78.6 & 79.3 & 3.93 \\
            8  & 72.3 & 78.9 & 79.5 & 5.43 \\
            10 & 72.7 & \textbf{79.5} & \textbf{80.1} & 7.94 \\
            \hline
        \end{tabular}}
    \end{minipage}
    \vspace{-0.3cm}
\end{table*}

\begin{figure*}[t]
  \centering
  \includegraphics[width=1\linewidth]{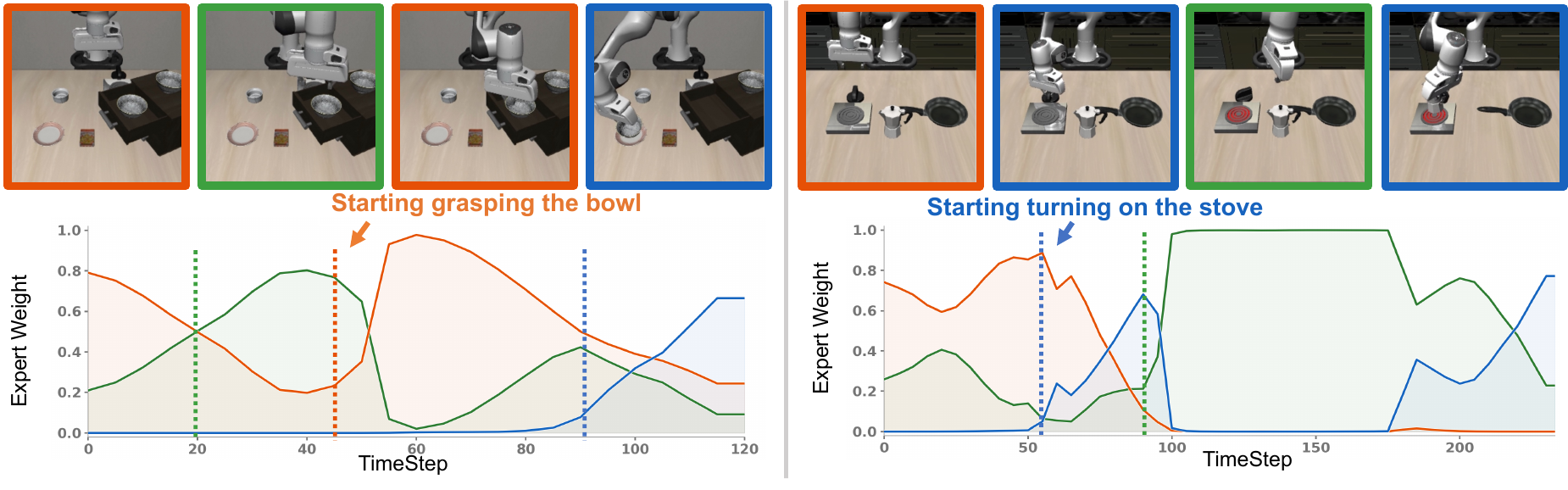}
  \vspace{-0.5cm}
  \caption{Visualization of expert activation weights over time. The gating coefficients dynamically shift and exhibit pronounced peaks during critical action phases such as \textit{grasping a bowl} or \textit{turning on a stove}, demonstrating that the EGTS module successfully drives individual experts to implicitly specialize in distinct, semantically meaningful motion primitives.}
  \label{visualization}
  \vspace{-0.33cm}
\end{figure*}

\noindent\textbf{Impact of Experts.} We further investigate the sensitivity of our framework to the number of experts ($N \in \{1, 2, 3, 5, 8, 10\}$), evaluating both task accuracy and computational efficiency in Table~\ref{ablation_expert}. As the number of experts increases, the model's ability to represent diverse skill primitives improves, leading to a steady rise in success rates. However, we observe diminishing returns beyond $N{=}5$; specifically, increasing the expert count to 10 provides only marginal accuracy improvement while substantially increasing inference time. Therefore, we adopt $N{=}5$ as the default configuration.

\noindent\textbf{Expert Activation Analysis.} To further validate the effectiveness of our implicit skill partitioning, we visualize the temporal evolution of expert routing weights during task execution. As illustrated in Figure~\ref{visualization}, the gating coefficients exhibit distinct, dynamic shifts that tightly align with critical motion transitions. Notably, when the robotic arm initiates significant semantic actions such as \textit{grasping a bowl} or \textit{turning on a stove}, the activation weights of specific experts demonstrate pronounced peaks. This temporal correlation confirms that the MoE routing mechanism successfully extracts and specializes in distinct latent atomic skills, even in the complete absence of explicit action labels.

\subsection{Simulated Evaluation}
To comprehensively evaluate the effectiveness of the proposed \textit{Skill-Based Memory} (SkillMemo) framework, we conduct experiments under both DP and VLA backbones across various widely used benchmarks: LIBERO, Push-T, UR3 Block Push, and Franka Kitchen.

\noindent\textbf{Benchmarks with DP backbone.}  
Follow DP~\citep{chi2023diffusion}, we evaluate our framework on three standard continuous-control environments: Push-T, UR3 Block Push, and Franka Kitchen. Table~\ref{dp_sota} presents a comparative analysis of SkillMemo against various state-of-the-art baselines, including Diffusion Policy (DP)~\citep{chi2023diffusion}, Sparse Diffusion Policy (SDP)~\citep{wang2024sparse}, Consistency Policy (CP)~\citep{prasad2024consistency}, IMLE Policy~\citep{rana2025imle}, and STEP~\citep{li2026step}. While existing methods fundamentally lack explicit temporal memory and struggle with fine-grained skill extraction, severely limiting their high-level task compositionality. In contrast, by structurally storing and dynamically retrieving latent atomic skills, SkillMemo consistently outperforms all baselines across every benchmark, achieving the highest average success rate of 64.6\%, where surpassing DP by 3.4\% and IMLE Policy by 1.8\%.

\begin{table*}[t]
    \centering
    \small
    \renewcommand\arraystretch{1.3}
    \definecolor{ourshighlight}{HTML}{FCE4D6}
    \begin{minipage}[t]{0.48\textwidth}
        \centering
        \caption{Success rates compared to state-of-the-art Diffusion Policy(DP) baselines across various simulated benchmarks.}
        \label{dp_sota}
        \vspace{-0.1cm}
        \resizebox{\linewidth}{!}{%
        \begin{tabular}{c|ccc|c}
            \hline
            Model & Push-T & BlockPush & Kitchen & Avg.\\
            \hline
            DP~\citep{chi2023diffusion} & 52.9 & 73.6 & 57.1 & 61.2 \\
            SDP~\citep{wang2024sparse} & 53.5 & 72.9 & 56.3 & 60.9 \\
            CP~\citep{prasad2024consistency} & 53.2 & 73.2 & 56.0 & 60.8 \\
            IMLE Policy~\citep{rana2025imle} & 53.7 & 77.2 & 57.5 & 62.8 \\
            STEP~\citep{li2026step} & 49.7 & 76.9 & 58.1 & 61.6 \\
            \rowcolor{ourshighlight} 
            \textbf{SkillMemo} & 55.2 & 78.1 & 60.6 & 64.6 \\
            \hline
        \end{tabular}}
    \end{minipage}%
    \hfill
    \begin{minipage}[t]{0.48\textwidth}
        \centering
        \caption{Zero-shot generalization to unseen task configurations. Models trained on specific LIBERO suites are evaluated cross-suite.}
        \label{sim_unseen}
        \vspace{-0.1cm}
        \resizebox{\linewidth}{!}{%
        \begin{tabular}{c|c|ccc}
            \hline
            Dataset & Method & Goal & Spatial & Object\\
            \hline
            \multirow{2}{*}{LIBERO-Goal} & w/o memory & 68.7 & 71.3 & 72.5 \\
            & w memory & 73.4 & 75.6 & 79.1 \\
            \hline
            \multirow{2}{*}{LIBERO-Spatial} & w/o memory & 70.3 & 72.1 & 73.9 \\
            & w memory & 72.0 & 80.2 & 78.2 \\
            \hline
            \multirow{2}{*}{LIBERO-Object} & w/o memory & 69.5 & 70.8 & 75.7 \\
            & w memory & 71.6 & 73.4 & 79.8 \\
            \hline
        \end{tabular}}
    \end{minipage}
    \vspace{-0.3cm}
\end{table*}


\noindent\textbf{Benchmarks with VLA backbone.} 
Following OpenVLA~\citep{kim2024openvla}, we evaluate our framework on the LIBERO benchmark~\citep{liu2023libero} utilizing a 7-DoF Franka Panda robot across different suites: \textit{Goal}, \textit{Spatial}, \textit{Object}, and \textit{Long}. Each suite consists of 10 distinct manipulation tasks, while LIBERO-Long focuses on long-horizon tasks. We follow $\pi_{0.5}$~\citep{intelligence2025pi_} which training single unified policy across all tasks for 30k steps. Validation reports the average success rate over 50 rollouts per task.

\begin{table*}[t]
    \centering
    \small
    \renewcommand\arraystretch{1.3} 
    \setlength{\tabcolsep}{12pt}   
    
    \definecolor{ourshighlight}{HTML}{FCE4D6}
    
    \caption{Success rates on the LIBERO benchmark. Our proposed SkillMemo framework consistently enhances the performance of various state-of-the-art VLA backbones across different scales.}
    \label{tab:main_libero_sota}
    \vspace{-0.1cm}
    
    \resizebox{\linewidth}{!}{%
    \begin{tabular}{l | c | c c c c | c}
        \toprule
        \multirow{2}{*}{\textbf{Method}} & \multirow{2}{*}{\textbf{Params}} & \multicolumn{4}{c|}{\textbf{LIBERO Benchmark (\%)} $\uparrow$} & \multirow{2}{*}{\textbf{Average} $\uparrow$} \\
        \cmidrule{3-6}
        & & \textbf{Goal} & \textbf{Spatial} & \textbf{Object} & \textbf{Long} & \\
        \midrule
        OpenVLA~\citep{kim2024openvla} & 7.0B & 78.0 & 85.0 & 86.8 & 54.0 & 76.0 \\
        TriVLA~\citep{liu2025trivla} & 3.4B & 89.8 & 91.2 & 93.8 & 73.2 & 87.0 \\
        CogACT~\citep{li2024cogact} & 7.6B & 90.2 & 97.2 & 98.0 & 88.8 & 93.2 \\
        $\pi_0$~\citep{black2024pi_0} & 3.3B & 95.8 & 96.8 & \textbf{98.8} & 85.2 & 94.2 \\
        MemoryVLA~\citep{shi2025memoryvla} & 7.3B & 96.4 & 98.4 & 98.4 & 93.4 & 96.5 \\
        \midrule
        UniAct~\citep{zheng2025universal} & 0.5B & 68.7 & 72.1 & 75.7 & 51.4 & 67.2 \\
        \rowcolor{ourshighlight} 
        \textbf{UniAct-SkillMemo (Ours)} & \textbf{0.6B} & \textbf{73.4} & \textbf{80.2} & \textbf{79.8} & \textbf{57.2} & \textbf{72.7} \\
        \midrule
        UniVLA~\citep{bu2025univla} & 8.5B & 91.8 & 96.5 & 95.6 & 92.0 & 93.9 \\
        \rowcolor{ourshighlight} 
        \textbf{UniVLA-SkillMemo (Ours)} & \textbf{8.9B} & \textbf{94.5} & \textbf{98.1} & \textbf{97.8} & \textbf{93.2} & \textbf{95.9} \\
        \midrule
        $\pi_{0.5}$~\citep{intelligence2025pi_} & 3.3B & 98.0 & 98.8 & 98.2 & 92.4 & 96.8 \\
        \rowcolor{ourshighlight} 
        \textbf{$\pi_{0.5}$-SkillMemo (Ours)} & \textbf{3.6B} & \textbf{99.0} & \textbf{99.4} & 98.2 & \textbf{95.4} & \textbf{98.0} \\
        \bottomrule
    \end{tabular}}
    \vspace{-0.45cm}
\end{table*}

\begin{table*}[t]
    \centering
    \small
    \caption{Single-task performance in real world. Success rates (\%) are reported over 40 trials per task.}
    \renewcommand\arraystretch{1.3}
    \definecolor{ourshighlight}{HTML}{FCE4D6}
    \vspace{-0.1cm}
    \resizebox{\linewidth}{!}{%
    \begin{tabular}{c|ccccc} 
        \hline
         \makecell{Model} & \makecell{Strawberry in Bowl} & \makecell{Lemon on Plate} & \makecell{Corn in Pot} & \makecell{Butter in Pot} & \makecell{Chip Bucket on Table} \\
        \hline
        Diffusion Policy & 77.5 & 77.5 & 82.5 & 62.5 & 67.5 \\
        \rowcolor{ourshighlight} 
        SkillMemo & 82.5 & 80.0 & 90.0 & 75.0 & 77.5 \\
        \hline
    \end{tabular}}
    \label{tab:real_single}
    \vspace{-0.2cm}
\end{table*}


Table~\ref{tab:main_libero_sota} compares SkillMemo against cutting-edge VLA architectures on the LIBERO benchmark. While recent approaches like MemoryVLA~\citep{shi2025memoryvla} leverage holistic visual-semantic memory, they lack fine-grained behavioral priors for continuous action execution. Furthermore, even highly optimized foundation models such as $\pi_0$~\citep{black2024pi_0} and $\pi_{0.5}$~\citep{intelligence2025pi_} lack dedicated mechanisms to dynamically retrieve and compose historical skills. By structurally integrating latent atomic skills into a dynamic memory bank, SkillMemo consistently enhances these strong backbones. Notably, $\pi_{0.5}$-SkillMemo establishes average success rate of 98.0\%, yielding a 1.2\% (98.0\% vs.\ 96.8\%) improvement over $\pi_{0.5}$ and decisively surpassing MemoryVLA by 1.5\% (98.0\% vs.\ 96.5\%), demonstrating that skill-level memory is crucial for maximizing VLA performance.

\noindent\textbf{Unseen task generalization.}  
To further evaluate the generalization ability of the learned skills, we conduct a zero-shot cross-suite evaluation where models trained on the \textit{Goal}, \textit{Spatial}, and \textit{Object} subsets respectively are directly evaluated on the other subsets. As shown in Table~\ref{sim_unseen}, the SkillMemo framework trained on \textit{Goal} achieves 75.6\% on unseen \textit{Spatial} suite, which even outperforms the in-distribution baseline with 3.5\% (75.6\% vs.\ 72.1\%) trained explicitly on \textit{Spatial}. This result strongly supports our hypothesis that the memory bank enables compositional reuse of skill primitives, allowing the model to generalize beyond the training distribution by recombining known skills.

\begin{table*}[t]
    \vspace{0.1cm}
    \centering
    \renewcommand\arraystretch{1.15}
    \small
    \caption{Real-world generalization to unseen tasks. Marks with \Checkmark indicates the source tasks used for training. We evaluate the models on unseen compositional tasks to test skill transferability. Success rates (\%) are reported over 40 trials.}
    \vspace{-0.1cm}
    \resizebox{\linewidth}{!}{%
    \begin{tabular}{c|ccc|c|c} 
        \hline
        \makecell{Unseen Tasks} & \makecell{Strawberry in Bowl} & \makecell{Lemon on Plate} & \makecell{Butter in Pot} & \makecell{Diffusion Policy} & \makecell{SkillMemo} \\
        \hline
        Strawberry in Bowl & \Checkmark & \Checkmark & & 77.5 & 82.5 \\
        Lemon on Plate & \Checkmark & \Checkmark & & 77.5 & 80.0 \\
        Strawberry on Plate & \Checkmark & \Checkmark & & 65.0 & 75.0 \\
        Lemon in Bowl & \Checkmark & \Checkmark & & 62.5 & 75.0 \\
        \hline
        Strawberry in Pot & \Checkmark & \Checkmark & \Checkmark & 60.0 & 70.0 \\
        Lemon in Pot & \Checkmark & \Checkmark & \Checkmark & 57.5 & 72.5 \\
        Butter in Bowl & \Checkmark & \Checkmark & \Checkmark & 65.0 & 70.0 \\
        Butter on Plate & \Checkmark & \Checkmark & \Checkmark & 60.0 & 67.5 \\
        \hline
    \end{tabular}}
    \label{tab:real_unseen}
    \vspace{-0.25cm}
\end{table*}

\subsection{Real-World Evaluation}

\noindent\textbf{Single-Task Evaluation.} 
Following the hardware setup described in the Implementation Details, we deploy the policies on a real-world UR5e robotic manipulator. We curate a suite of five contact-rich manipulation tasks involving diverse objects and semantic goals: \textit{Put the Strawberry into the Bowl}, \textit{Place the Lemon on the Plate}, \textit{Put the Corn into the Pot}, \textit{Grasp the Butter in the Pot}, and \textit{Place the Chip Bucket on the Table}. For each task, we collect a dataset of 50 expert demonstrations and evaluate using 40 independent rollouts. As shown in Table~\ref{tab:real_single}, SkillMemo consistently outperforms DP across all five tasks, with particularly notable improvements on more challenging tasks such as \textit{Corn in Pot} (+7.5\%) and \textit{Butter in Pot} (+12.5\%).

\noindent\textbf{Unseen Task Generalization.}  
To investigate the capability of SkillMemo to generalize to unseen object-receptacle combinations, we design a compositional evaluation protocol. Models are trained on a limited subset of source tasks and subsequently evaluated on novel pairings. As detailed in Table~\ref{tab:real_unseen}, we establish two generalization settings. In the first scenario, models are co-trained on \textit{Strawberry in Bowl} and \textit{Lemon on Plate}, and then tested on the unseen crossover combinations. In the second scenario, we extend the training set with \textit{Butter in Pot} and evaluate generalization to more complex unseen targets. The results indicate that SkillMemo significantly outperforms DP in these compositional scenarios, improving \textit{Strawberry on Plate} from 65.0\% to 75.0\%. This confirms that our framework effectively retrieves and fuses atomic skills from disparate training experiences to facilitate robust compositional generalization on unseen task configurations.






\section{Conclusion}
We presented SkillMemo, a skill memory framework for compositional embodied manipulation that addresses the limitations of existing visuomotor models in capturing and reusing structured behavioral patterns. By coupling expert-guided trajectory segmentation with a skill-level episodic memory architecture, SkillMemo enables the explicit decomposition, storage, retrieval, and composition of atomic skill primitives. Extensive experiments on both simulation benchmarks and real-world robotic tasks demonstrate that SkillMemo consistently enhances the performance of diverse backbone architectures while exhibiting zero-shot compositional generalization to unseen task compositions.

\clearpage


\bibliography{egbib}
\clearpage
\appendix

\section{Extended Implementation Details}
\label{app:implementation}

\subsection{Simulation Setup}
For simulation experiments, we conduct evaluations on the widely utilized LIBERO benchmark~\citep{liu2023libero}. LIBERO comprises a collection of long-horizon and language-conditioned robotic manipulation tasks, designed to test generalization to compositional and unseen task instructions. All models are trained on 8 NVIDIA A6000 GPUs with PyTorch. We use 32 samples per GPU for a global batch size of 256 and adopt AdamW with a learning rate of $2\times10^{-5}$. The policy takes as input a single RGB image observation and a language instruction, and outputs a 7-dimensional continuous action (6-DoF end-effector pose plus gripper command). The MoE module consists of $N{=}5$ experts by default. The memory bank stores up to $M_{\text{max}}{=}1000$ skill entries.

\subsection{Real-World Setup}
\begin{figure}[htbp]
    \centering
    \includegraphics[width=0.55\linewidth]{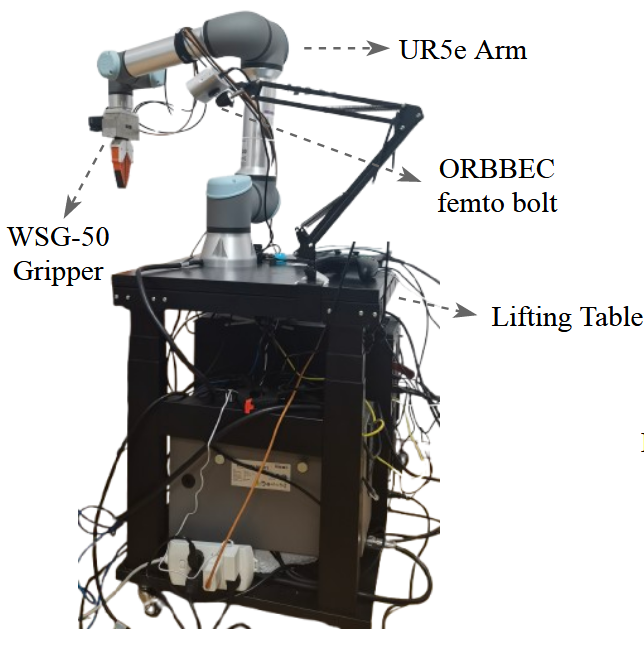}
    \caption{Hardware setup for real-world experiments, featuring a UR5e manipulator, WSG-50 gripper, ORBBEC Femto Bolt camera, and a mobile lifting table.}
    \label{fig:equipment}
\end{figure}

For real-world validation, we deploy the DP model on a single 7-DoF UR5e robotic manipulator integrated with a Weiss WSG-50 parallel-jaw gripper, as illustrated in Figure~\ref{fig:equipment}. An ORBBEC Femto Bolt RGB-D camera is mounted on a flexible boom arm in a side-view configuration to capture visual observations. The robotic arm is anchored to a height-adjustable lifting table with a mobile base, which facilitates the evaluation of policy robustness across varying viewpoints and mounting heights. The action space is defined as a 6-dimensional vector, comprising 3D end-effector translation, 3D rotation (Euler angles), and a binary gripper command. We collect 50 expert demonstrations per task through kinesthetic teaching.

\subsection{Qualitative Results in the Real World}
\begin{figure*}[htbp] 
    \centering
    \includegraphics[width=\linewidth]{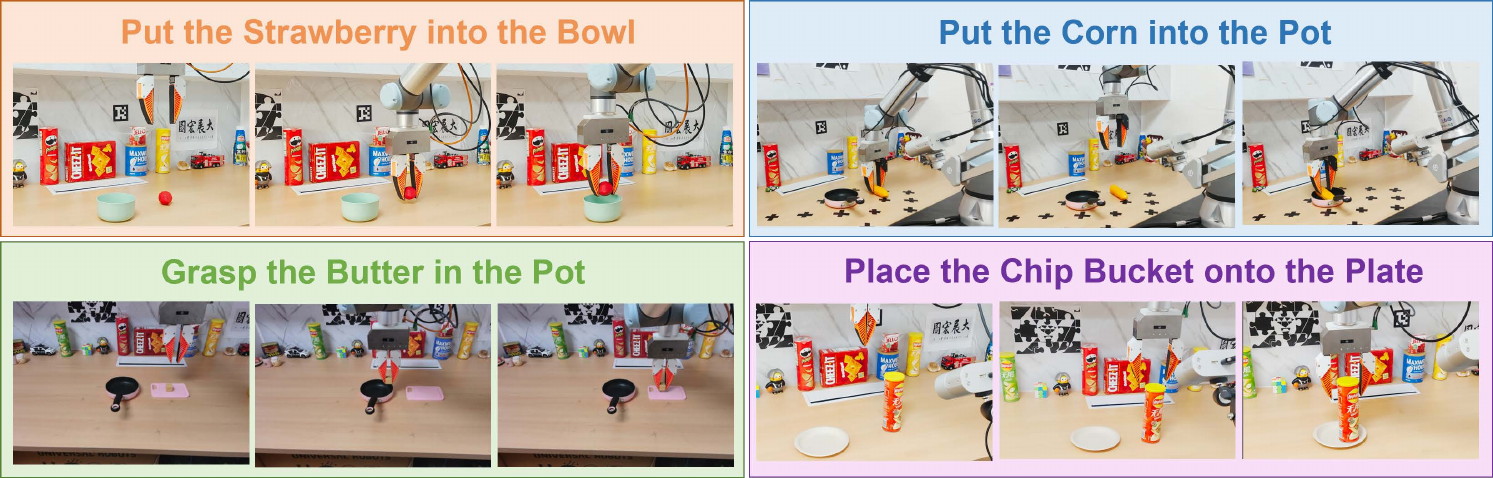}
    \caption{Visualization of successful rollouts for real-world manipulation tasks using the SkillMemo framework. The continuous frames illustrate the UR5e robot effectively executing diverse compositional tasks, demonstrating accurate skill retrieval, robust grasping, and precise placement across different object-receptacle combinations.}
    \label{fig:real_tasks}
\end{figure*}

To further demonstrate the effectiveness and robustness of our proposed SkillMemo framework in physical environments, we provide qualitative visualizations of the real-world task executions. As shown in Figure~\ref{fig:real_tasks}, we visualize the continuous execution frames of four distinct compositional manipulation tasks: \textit{Place the Strawberry into the Bowl}, \textit{Place the Corn into the Pot}, \textit{Put the Butter into the Pot}, and \textit{Put the Chip Bucket onto the Table}. 

The visualizations confirm that by retrieving and composing latent atomic skills from the memory bank, the UR5e manipulator can smoothly and accurately handle varied objects and receptacles, exhibiting precise grasping and placing behaviors even under complex real-world dynamics.

\end{document}